%% file: root.tex
\documentclass[letterpaper, 10 pt, conference]{ieeeconf}  

\IEEEoverridecommandlockouts                              

\usepackage{url} 

\usepackage[acronym,nonumberlist]{glossaries}
\include{acronyms}

\usepackage{graphicx}

\usepackage{tikz,pgfplots,pgfplotstable}
\pgfplotsset{compat=1.8}
\usepackage{calc}

\usepackage{tikzscale}

\usepackage{subcaption}

\usepackage{floatrow}

\usepackage{siunitx} 

\usepackage{listings}
\usepackage{atbegshi}

\title{\LARGE \bf
3D Coverage Path Planning for Efficient Construction Progress Monitoring
}

\author{Katrin Becker, Martin Oehler and Oskar von Stryk
\thanks{All authors are with the Simulation, Systems Optimization and Robotics Group, Technical University of Darmstadt, Germany.\linebreak
{\tt\small \{kbecker, oehler, stryk\}@sim.tu-darmstadt.de}}
\thanks{Research presented in this paper has been supported by Nexplore within the AICO Collaboration Lab.
\newline 978-1-6654-5680-7/22/\$31.00 \textcopyright 2022 IEEE}%
}

\AtBeginShipoutFirst{\raisebox{1cm}{\parbox{0.9\textwidth}{\centering\normalsize Preprint of the paper which appeared in:\\
            \large IEEE International Symposium on Safety, Security, and Rescue Robotics (SSRR) 2022}}}

\begin{document}

\maketitle
\thispagestyle{empty}
\pagestyle{empty}

\begin{abstract}
On construction sites, progress must be monitored continuously to ensure that the current state corresponds to the planned state in order to increase efficiency, safety and detect construction defects at an early stage.
Autonomous mobile robots can document the state of construction with high data quality and consistency.
However, finding a path that fully covers the construction site is a challenging task as it can be large, slowly changing over time, and contain dynamic objects.
Existing approaches are either exploration approaches that require a long time to explore the entire building, object scanning approaches that are not suitable for large and complex buildings, or planning approaches that only consider 2D coverage.

In this paper, we present a novel approach for planning an efficient 3D path for progress monitoring on large construction sites with multiple levels. By making use of an existing 3D model we ensure that all surfaces of the building are covered by the sensor payload such as a 360-degree camera or a lidar. This enables the consistent and reliable monitoring of construction site progress with an autonomous ground robot. We demonstrate the effectiveness of the proposed planner on an artificial and a real building model, showing that much shorter paths and better coverage are achieved than with a traditional exploration planner.
\end{abstract}

\section{INTRODUCTION}

In modern construction, each stage of construction is planned as a digital model in the process of \gls{bim}. Continuously monitoring the progress of the construction ensures that the current state matches the original plan. In this way, construction defects can be detected at an early stage, improving safety and productivity during construction. The state of the construction site is documented in a ``digital twin" that accumulates data over time such as 360-camera images or lidar point clouds.
In order to improve data quality, it is necessary to have a monitoring plan showing exactly where to record the required data. This leads to a more complete model and more consistent and comparable data.
Data for such a model could be manually captured by humans following the plan and taking photos and scans of the environment. However, humans tend to make errors when working on long repetitive tasks, leading to incomplete data.
Therefore, this task can be automated using autonomous mobile robots.
They can precisely follow the plan and capture data with high consistency. Ground robots are well suited for construction sites because they can safely navigate inside buildings and carry heavy sensor payloads.

Creating a monitoring plan that covers all surfaces of a large building is a challenging task. Existing object scanning approaches \cite{cunningham, daudelin}  are often limited to small objects and scenes, and do not allow for the robot to be inside the object or scene. Exploration approaches \cite{exp_transf, tare} on the other hand do not use prior information about the environment and take a long time to explore the complete building. There are approaches to efficiently scan existing buildings \cite{chen}, but they use a floor plan for planning and thus can only ensure 2D coverage.

\begin{figure}[tpb]
\centering
\includegraphics[width=\linewidth]{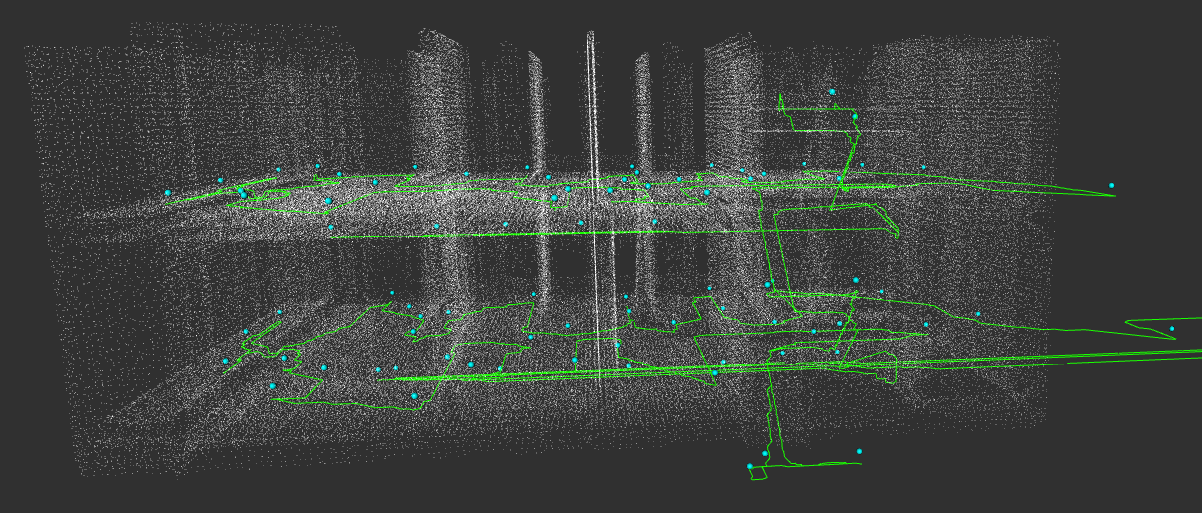}
\caption{Path planned on a real building model over multiple levels for a smaller target model (white point cloud). The dots are the selected viewpoints, the green line is the planned path. It has 84 waypoints and is \SI{520.77}{\meter} long.}
\label{fig:precom_res_large_target_multilevel}
\end{figure}

We propose a novel approach to compute a time-optimal 3D path for monitoring large construction sites with multiple levels (Fig. \ref{fig:precom_res_large_target_multilevel}). By utilizing the 3D building model as prior information, we create an efficient monitoring plan for an autonomous ground robot for covering all surfaces of the construction site with a sensor payload. The approach supports various 3D sensors such as 360-degree cameras and lidar. We demonstrate the effectiveness of the proposed planner on an artificial and a real building model. In simulation, we compare our approach to an exploration planner \cite{exp_transf} showing that we achieve much shorter paths and a better coverage with a lidar scanner.

\begin{figure*}[thpb]
      \centering
      \includegraphics[width=\textwidth]{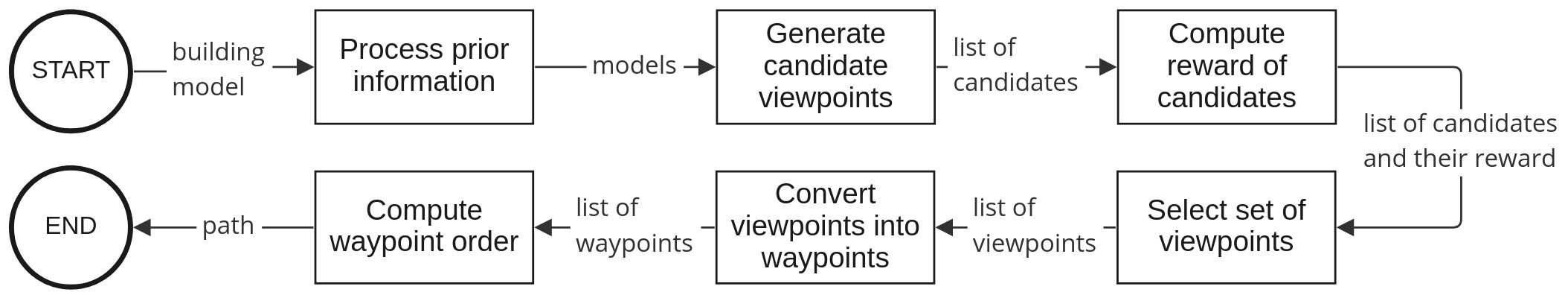}
      \caption{Planning pipeline}
      \label{fig:precom_overview}
\end{figure*}

\section{Related Work} \label{sec:rel_work}

Planning a path that covers all surfaces of an environment combines aspects from various research on different tasks, among others, object scanning, exploration, and the scanning of existing buildings which is often not automated with robots, but done with e.g. terrestrial laser scanners. All the mentioned tasks are mostly solved using \gls{nbv} planning. A common approach for \gls{nbv} planning is to first generate a set of candidate viewpoints based on different criteria. Afterwards, these candidates are ranked by their reward. Finally, either the best candidate is chosen as \gls{nbv} or a set of candidates is selected as viewpoints.

In the approach proposed by Cunningham-Nelson et al. \cite{cunningham}, a fixed object is scanned with a mobile manipulator. It starts with no information about the scene except the object position and first estimates a prior model by navigating to waypoints in a cone shape. For each generated candidate viewpoint, it is checked how many clusters from the prior model are in the sensor's field of view. A subset of candidate viewpoints is selected by using a greedy approach which always selects the candidate with the most seen clusters. Afterwards, one or more viewing angles are generated from which all associated clusters are visible.

Daudelin and Campbell \cite{daudelin} focus on the scanning of small object scenes of unknown size. The \gls{nbv} is computed online based on frontier cells, which are unknown cells that border empty and occupied cells and which have the highest probability of belonging to the object. The candidate positions are generated dynamically and evaluated for as many orientations as required to cover all directions of view.

Cao et al. \cite{tare} propose an exploration algorithm for large environments. The goal is to find the shortest path that covers all uncovered surfaces in the local search space. The uniformly sampled candidate viewpoints are rewarded with the area of surfaces they can cover. According to the reward of the viewpoints, a minimum set of viewpoints is chosen probabilistically. Afterwards, a path based on the viewpoints is computed by solving a \acrfull{tsp}. These two steps, the viewpoint selection and the path planning, are repeated a given number of times and the overall best solution, i.e. the one with the lowest path costs, is used. Therefore, the method optimizes the entire exploration path, instead of only maximizing the instant reward.

Based on a given 2D floor plan, Chen et al. \cite{chen} plan where to place a terrestrial laser scanner to create a 3D model of an existing building. Since here the lines that form the walls in the floor plan are used, only 2D coverage can be ensured. In order to rate the generated candidates, a sweep-ray algorithm is used to determine the visible line segments with respect to a minimum and maximum viewing distance. For selecting the minimum set of viewpoints, a greedy best-first search, a greedy algorithm with a backtracking process, and a simulated annealing algorithm are used.

\label{rel_work:current_exploration}
An exploration algorithm is used for the comparison with the proposed approach (cf. Section \ref{eval:comp_exp}). It is based on the work proposed by Wirth and Pellenz \cite{exp_transf} and uses a 2D occupancy grid. Here, the next exploration target is selected based on the distance from frontier cells, in this case, free cells adjacent to unknown cells, and the cost of the path to it based on the distance and safety of the route.

\section{Method}
We propose a \acrfull{nbv} planner to generate inspection routes for monitoring construction progress. In contrast to most of the work presented in Section \ref{sec:rel_work}, the planning process is based on a prior 3D model of the construction site and is performed offline in advance.

Fig. \ref{fig:precom_overview} shows an overview of our planning pipeline. First, we convert the given 3D model of the construction site to various representations used for planning. Using these models, candidate viewpoints for data recording are generated. For each candidate, a reward is computed that reflects the covered surface area. Based on the reward, a subset of viewpoints is selected that optimally covers all surfaces of the building. After converting the sensor viewpoints to robot waypoints, the shortest path that connects these waypoints is computed. The modular design of the planning pipeline allows to easily exchange steps if a better approach is found.

\subsection{Process prior information} \label{method:precomputations}

\begin{figure*}[thpb]
\centering
\begin{subfigure}[t]{.24\textwidth}
  \centering
  \includegraphics[width=\linewidth]{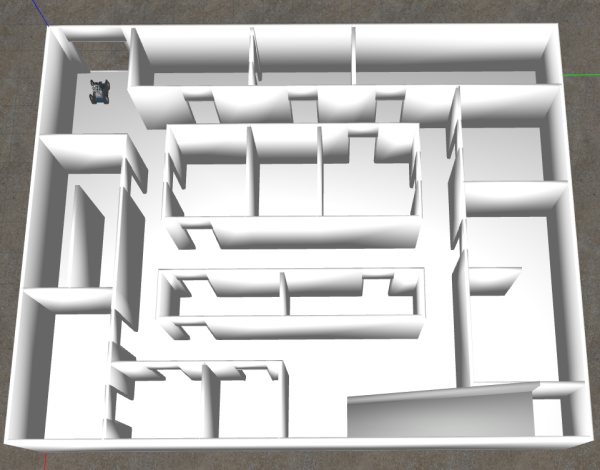}
  \caption{Artificial building model.}
  \label{fig:artificial_model}
\end{subfigure}\hfill%
\begin{subfigure}[t]{.24\textwidth}
  \centering
  \includegraphics[width=\linewidth]{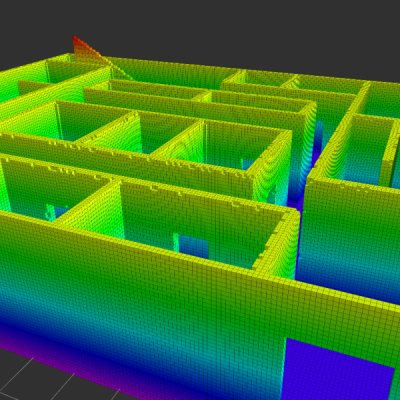}
  \caption{OctoMap}
  \label{fig:octomap}
\end{subfigure}\hfill%
\begin{subfigure}[t]{.24\textwidth}
  \centering
  \includegraphics[width=\linewidth]{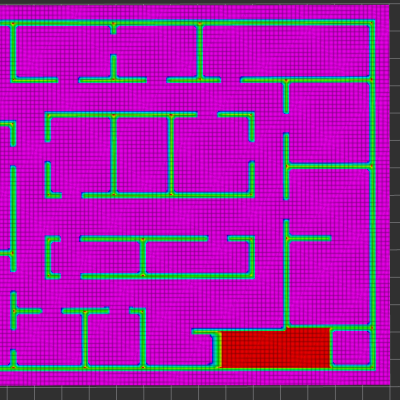}
  \caption{TSDF}
  \label{fig:sdf}
\end{subfigure}\hfill%
\begin{subfigure}[t]{.24\textwidth}
  \centering
  \includegraphics[width=\linewidth]{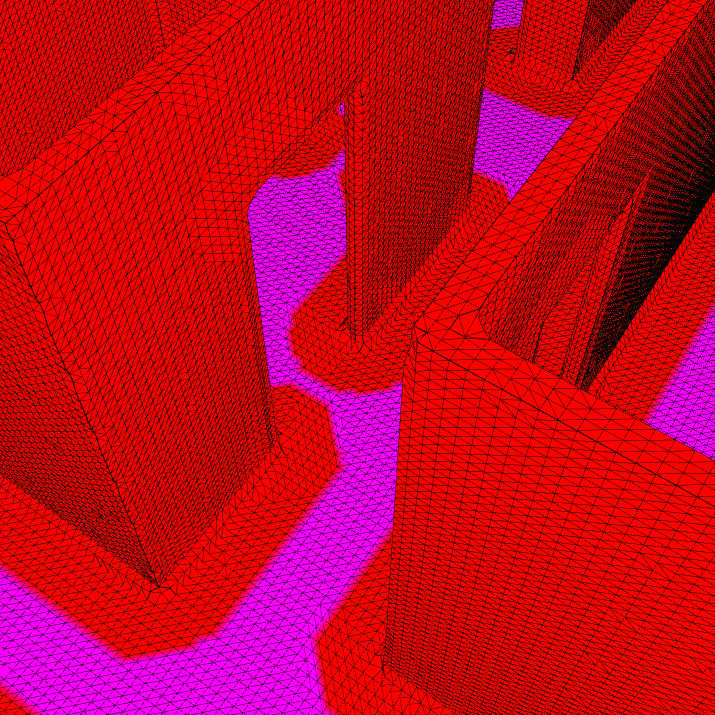}
  \caption{Mesh Map (violet: traversable space, red: non-traversable space)}
  \label{fig:mesh_map}
\end{subfigure}
\caption{Different model representations.}
\label{fig:models}
\end{figure*}

The prior information, i. e. the building mesh, needs to be present in two ways. First, as a complete model of the environment (Fig. \ref{fig:artificial_model}) used for path planning, and second, as a subset of the complete model, hereafter called the target model, which contains only the parts needed in the recorded data, e.g., only single rooms or also the complete model.

The complete model is converted into an OctoMap \cite{occupancy_grid, octomap} (Fig. \ref{fig:octomap}) and a \acrfull{sdf} \cite{esdf_tsdf} (Fig. \ref{fig:sdf}). The candidate generation and the waypoint order computation are performed using a mesh representation provided by the \textit{Mesh Navigation} \cite{mesh_navigation} (Fig. \ref{fig:mesh_map}). This mesh representation contains different layers indicating terrain traversability. We use a cost layer for local height differences and an inflation layer for lethal vertices in the first layer.
The target model is converted to a point cloud to form the target set, a set of points called target points, that is used to quantify the surface coverage.

\subsection{Generate candidate viewpoints}
The viewpoint candidate generation is split into the position and the orientation generation.
The candidate positions need to be generated equally distributed over the traversable space in order to handle multiple levels and height differences. Therefore, the mesh representation is iterated from a given start point and all reachable vertices, whose costs are below a given threshold and thus are seen as traversable, are used as candidates. In order to reduce this number, a uniformly distributed random sample is used.

Depending on the used sensor, multiple orientations are sampled for each candidate position in order to cover the complete environment. The number of equally distributed orientations $N$ per candidate position is given by
\begin{equation}
    N = \frac{n * 360\si{\degree}}{\mathit{fov}_h}
\end{equation}
where $n$ controls the overlap between the samples and $\mathit{fov}_h$ is the horizontal field of view of the sensor.

\subsection{Compute reward of candidates}
The reward of each candidate is given as the number of target points expected to be seen from its position according to the prior information.
In order to compute this information, the visibility is checked for each combination of candidate and target point. The general assumption is that a target point is visible if it is within the sensor's range and field of view and a clear line of sight exists between the candidate and target point.
In order to check if such a clear line of sight exists, ray casting can be used. However, since this is an expensive operation, other checks are performed beforehand to exclude invisible targets with as little computational effort as possible.

\begin{figure}
\floatbox[{\capbeside\thisfloatsetup{capbesideposition={right,top},capbesidewidth=0.4\linewidth}}]{figure}[\FBwidth]
    {\caption{Self filter mask for \textit{Asterix} (Fig. \ref{fig:asterix}). The visible points are green, the others red.}\label{fig:self-filter_mask}}
    {\includegraphics[width=\linewidth]{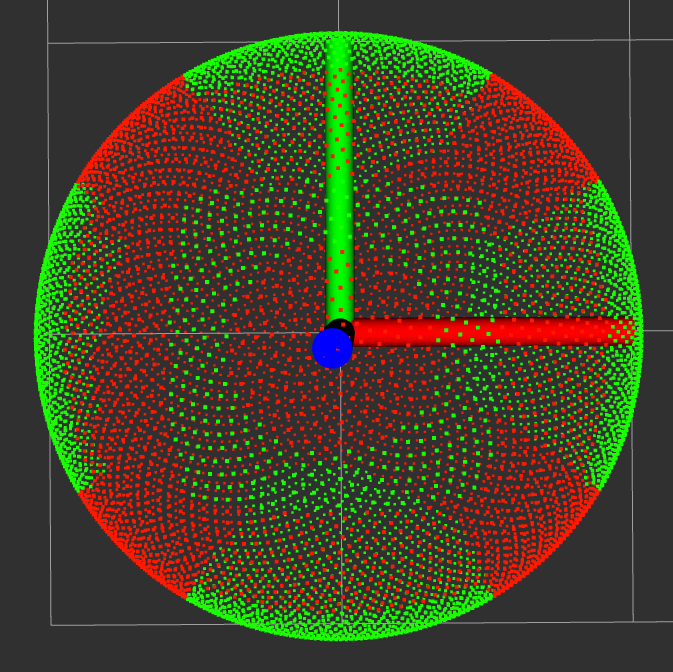}}
\end{figure}

The first visibility check ensures that the target point is within the sensor's range and field of view.
The next check rejects all the target points that cannot be seen since the view is blocked by the robot itself. For this, the \lstinline{robot_body_filter} \footnote{\url{http://wiki.ros.org/robot_body_filter}} is used. Since applying this filter for each candidate is costly, a mask (Fig. \ref{fig:self-filter_mask}) is precomputed and during the check for each target point, this point is mapped to the closest point on the mask. For the mask, a sampled unit sphere is used. The points on the mask are generated as spherical Fibonacci point sets \cite{spherical_fibonacci_mapping, fibonacci_lattice} and the mapping is performed as proposed by Keinert et al. \cite{spherical_fibonacci_mapping}.
In the third check, all target points located on the non-visible side of walls are excluded. To achieve this, we use the \acrshort{sdf} representation. If both, the direction from the target point to the candidate and the gradient at the target point have the same sign in each component, the target point is on the opposite side of a wall and cannot be seen. Finally, for the remaining points, it is checked if a clear line of sight exists using ray casting on the 3D occupancy grid map.

\subsection{Select set of viewpoints} \label{method:set_coverage_problem}
Next, the minimum number of candidate viewpoints that cover the target set as completely as possible must be selected. This corresponds to the ``set cover problem''. The solution to this problem is NP-complete, but there are heuristic solvers, three of which have been tested and compared here. The first one is a greedy approach which always selects the candidate with the highest reward. Each time a candidate is selected, the reward of the remaining ones is updated. But, since redundancies can occur with this approach if the covered set of one candidate is later completely covered by other candidates, the second approach introduces a backtracking step. The last approach is probabilistic. Instead of always selecting the best candidate, worse candidates can be selected based on an exponential distribution.

However, instead of selecting new candidates until the reward reaches 0, we introduce a minimum reward for a candidate to have in order to be selected. This minimum reward describes a trade-off between the number of viewpoints, on which the length of the path depends, and the coverage rate. The selection process is finished, when there are either no candidates or uncovered target points left, or when there is no candidate left whose reward is greater than the minimum reward.

\subsection{Compute waypoint order}
After the viewpoints have been selected, they are converted into waypoints by transforming them from the sensor frame to the robot's base link frame. To minimize the time for the execution of the plan, we find the shortest path connecting all waypoints.
Therefore, we use the length of the paths between every two waypoints as the cost and calculate the cost-optimal order of the waypoints. The paths are computed on the complete model using the \textit{Mesh Navigation} \cite{mesh_navigation}.
Choosing the best waypoint order with the lowest costs is a \acrfull{tsp} type problem. The problem here can be assumed as symmetric and metric. Solving the \acrshort{tsp} is NP-complete, but there exist heuristic algorithms. We use a simulated annealing algorithm \cite{sim_annealing} initialized with the minimum spanning tree approach \cite{cormen_introduction} to solve this problem.

\subsection{Execution of path}
For executing the path, the previously computed paths from the \acrshort{tsp} solution are used to approach the waypoints.
During execution, a 2D occupancy map is created. To react to dynamic obstacles, the precomputed path is continuously monitored by ensuring that the robot polygon for each pose of the planned path only contains free cells on the created map.
If an obstacle is found, only the blocked parts of the path are replanned, while all valid parts are preserved.
When a waypoint is reached, the data is recorded with the sensor payload, i.e. a 360-degree image or a lidar scan.

\section{Evaluation}
We evaluate our approach on two different models, a small artificial building (Fig. \ref{fig:artificial_model}) with an area of about \SI{183}{\square\meter} and a large model with a floor area of about \SI{1365}{\square\meter} which is part of a real building. The artificial building is constructed to contain difficulties that may also occur in real buildings, e.g. rooms with interior walls, multiple doors, and an unreachable room as well as a ramp in order to evaluate the planning on traversable height differences. The real building has two floors and multiple stairs.

\begin{figure}[thpb]
\centering
\begin{subfigure}{.48\linewidth}
  \centering
  \includegraphics[width=\linewidth]{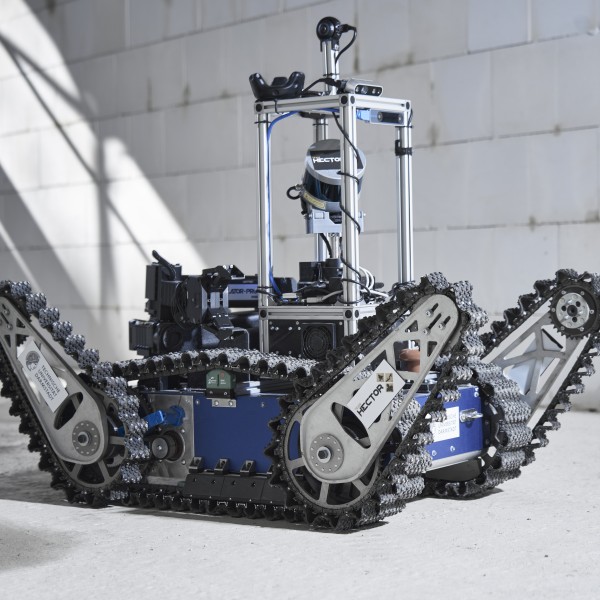}
  \caption{\textit{Asterix} \cite{asterix}}
  \label{fig:asterix}
\end{subfigure}\hfill%
\begin{subfigure}{.48\linewidth}
  \centering
  \includegraphics[width=\linewidth]{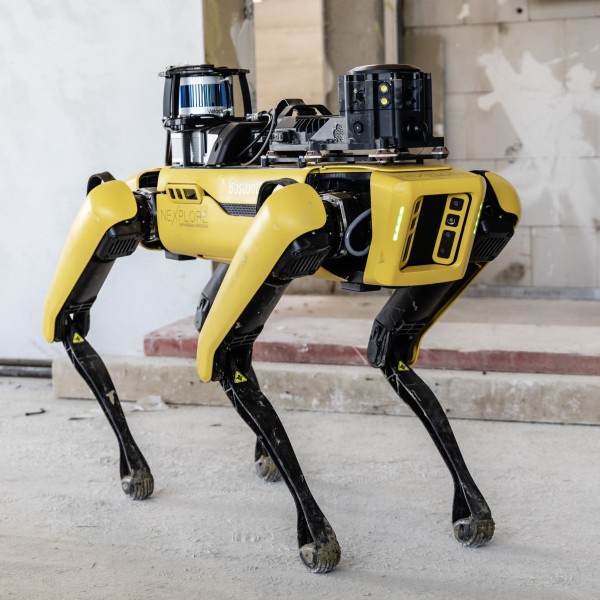}
  \caption{Spot (\copyright Nexplore)}
  \label{fig:spot}
\end{subfigure}
\caption{Robots used for evaluation. Both feature a 360-degree camera and a Velodyne VLP-16 lidar to scan the environment.}
\end{figure}

We demonstrate the generality of the approach by performing the evaluations on two different robots, \textit{Asterix} (Fig. \ref{fig:asterix}) and the Boston Dynamics Spot (Fig. \ref{fig:spot}). Both are highly mobile ground robots, each equipped with a sensor payload including a Velodyne VLP-16 lidar.

\subsection{Set cover problem}
When selecting the final viewpoints, a ``set cover problem" has to be solved (see Section \ref{method:set_coverage_problem}). We found the simple greedy approach to perform best in terms of computation time and number of selected viewpoints.
The variant without redundancies and the probabilistic approach could only marginally improve the result while having much longer computation times. Therefore, in the following experiments, the greedy approach has been used.

\begin{figure}[thpb]
      \centering
      \includegraphics[width=\linewidth]{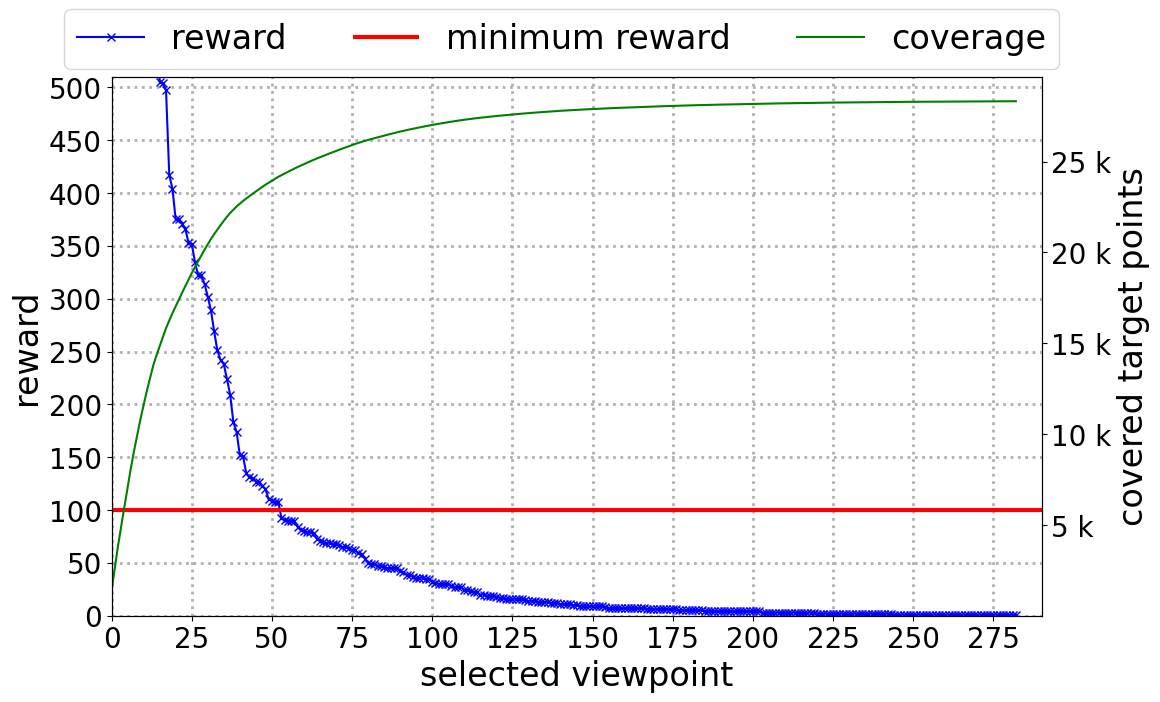}
      \caption{Reward of viewpoints in order of selection using the greedy approach. The viewpoints selected first are the best, the more points selected the less they contribute to coverage. Therefore, we chose a minimum reward of 100 (red line).}
      \label{fig:set_cov_cutoff}
\end{figure}

As described in Section \ref{method:set_coverage_problem}, we use a minimum reward to stop the selection of viewpoints early. In Fig. \ref{fig:set_cov_cutoff}, the reward of each newly selected viewpoint is shown as well as the number of covered target points. As can be seen, the reward of the newly selected viewpoints decreases very fast. Thus, without a specified minimum reward, many viewpoints would be selected without a considerable increase in coverage. For the following experiments, we set this value to 100, since there is a good balance between coverage and the number of viewpoints.

\subsection{Comparison to exploration planner}
\label{eval:comp_exp}

We compare our approach to an exploration planner that generates a map of an unknown environment. We use the algorithm proposed by Wirth and Pellenz \cite{exp_transf} for comparison. The driving distance has a high influence on the total time it takes to cover all surfaces of the building. Therefore, we compare the approaches in terms of total path length and coverage by comparing the acquired lidar point clouds.

\begin{figure}[tpb]
\centering
\begin{subfigure}{0.49\linewidth}
  \centering
  \includegraphics[width=\linewidth]{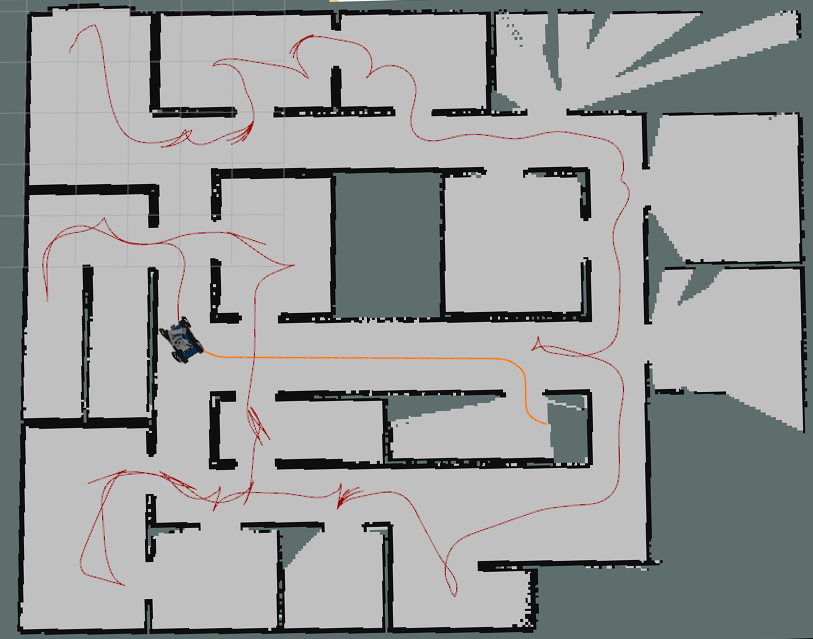}
  \caption{Exploration in process.}
  \label{fig:exploration_process}
\end{subfigure}\hfill
\begin{subfigure}{0.49\linewidth}
  \centering
  \includegraphics[width=\linewidth]{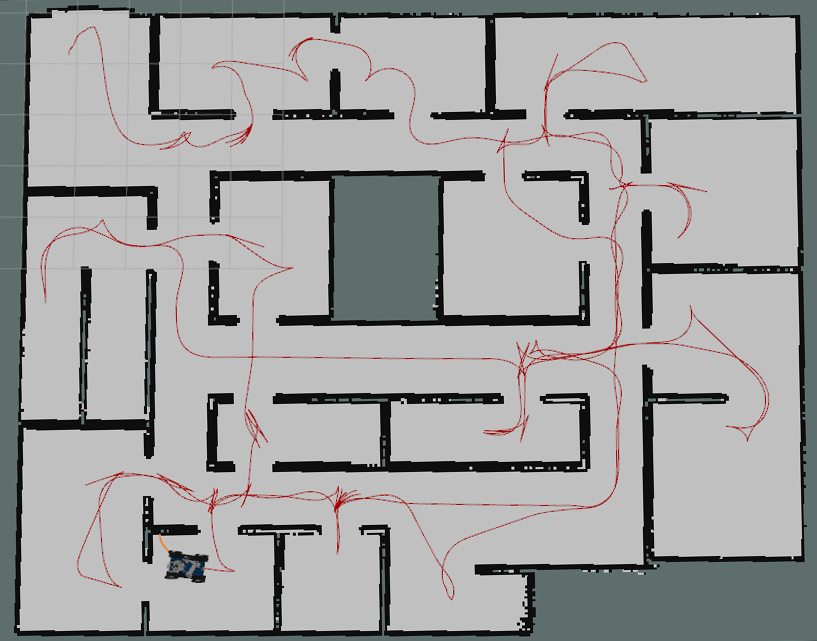}
  \caption{Finished exploration.}
  \label{fig:exploration_complete}
\end{subfigure}
\caption{Exploration on the artificial building model using \textit{Asterix}. The images contain the 2D occupancy grid map, the driven trajectory (dark red), the current robot position, and the path to the next exploration goal (orange).}
\label{fig:exploration_path_gridmap}
\end{figure}

\begin{figure}[thpb]
\centering
\includegraphics[width=\linewidth]{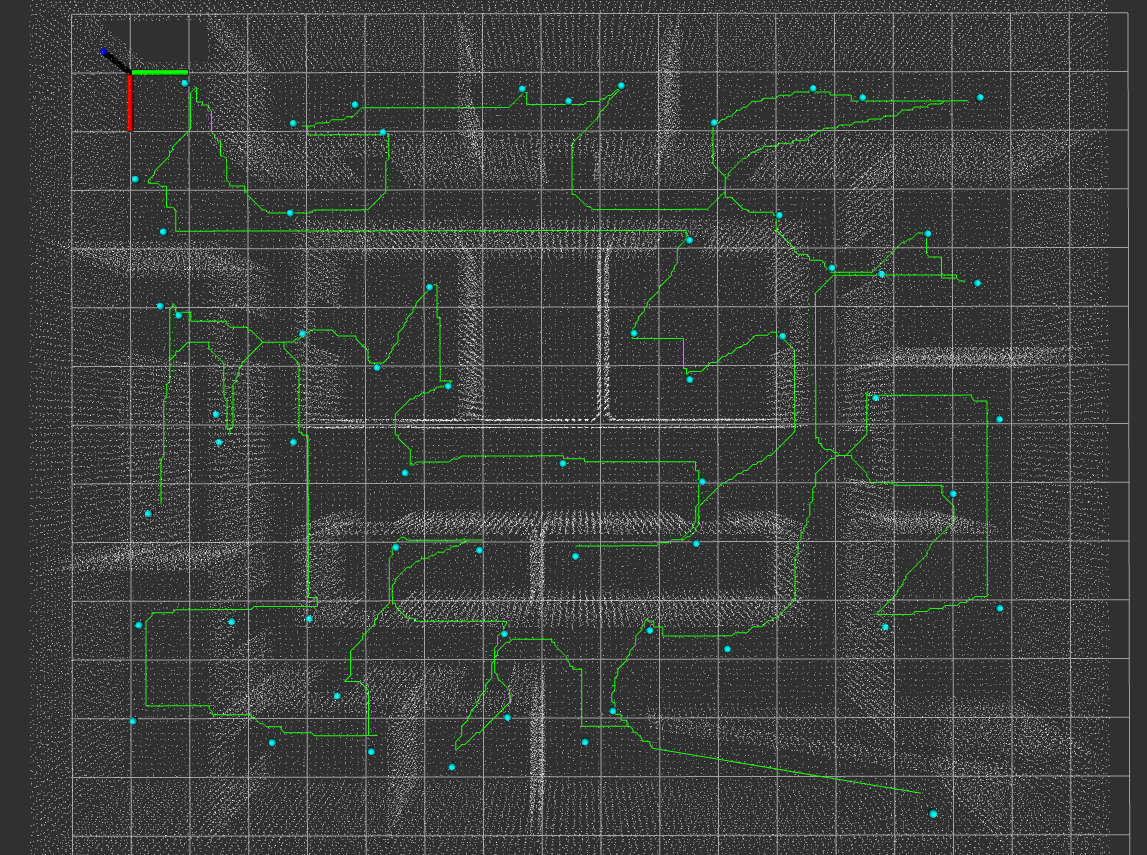}
\caption{Path planned on artificial building model using \textit{Asterix} with complete model as target model (white point cloud). The dots are the selected viewpoints, the green line is the planned path.}
\label{fig:precom_res_small_complete}
\end{figure}

The path of the exploration on the artificial model using \textit{Asterix} can be seen in Fig. \ref{fig:exploration_path_gridmap}. The driven path has a length of \SI{172}{\meter}. The ramp on the bottom side is missing because the planner cannot handle height differences. Due to the greedy approach, many small uncovered areas are remaining after the first pass of the building as can be seen in Fig. \ref{fig:exploration_process}. Returning to these uncovered areas leads to path redundancies.
In contrast, Fig. \ref{fig:precom_res_small_complete} contains the path planned with the presented approach. Here, the path is much shorter with only \SI{155.9}{\meter} and contains 62 waypoints. As can be seen, all reachable areas including the ramp contain viewpoints and the path has no unnecessary redundancies.

\begin{figure}[tpb]
\centering
\begin{subfigure}{\linewidth}
  \centering
  \includegraphics[width=\linewidth]{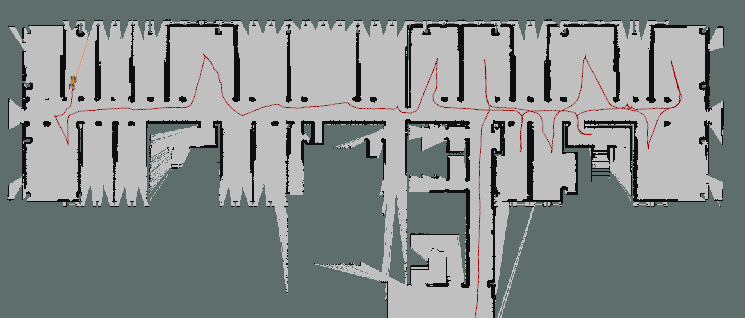}
  \caption{Partial exploration. The image contains the 2D occupancy grid map and the driven trajectory (dark red).}
  \label{fig:exploration_path_gridmap_spot_large}
\end{subfigure}\vspace{0.5cm}
\begin{subfigure}{\linewidth}
  \centering
  \includegraphics[width=\linewidth]{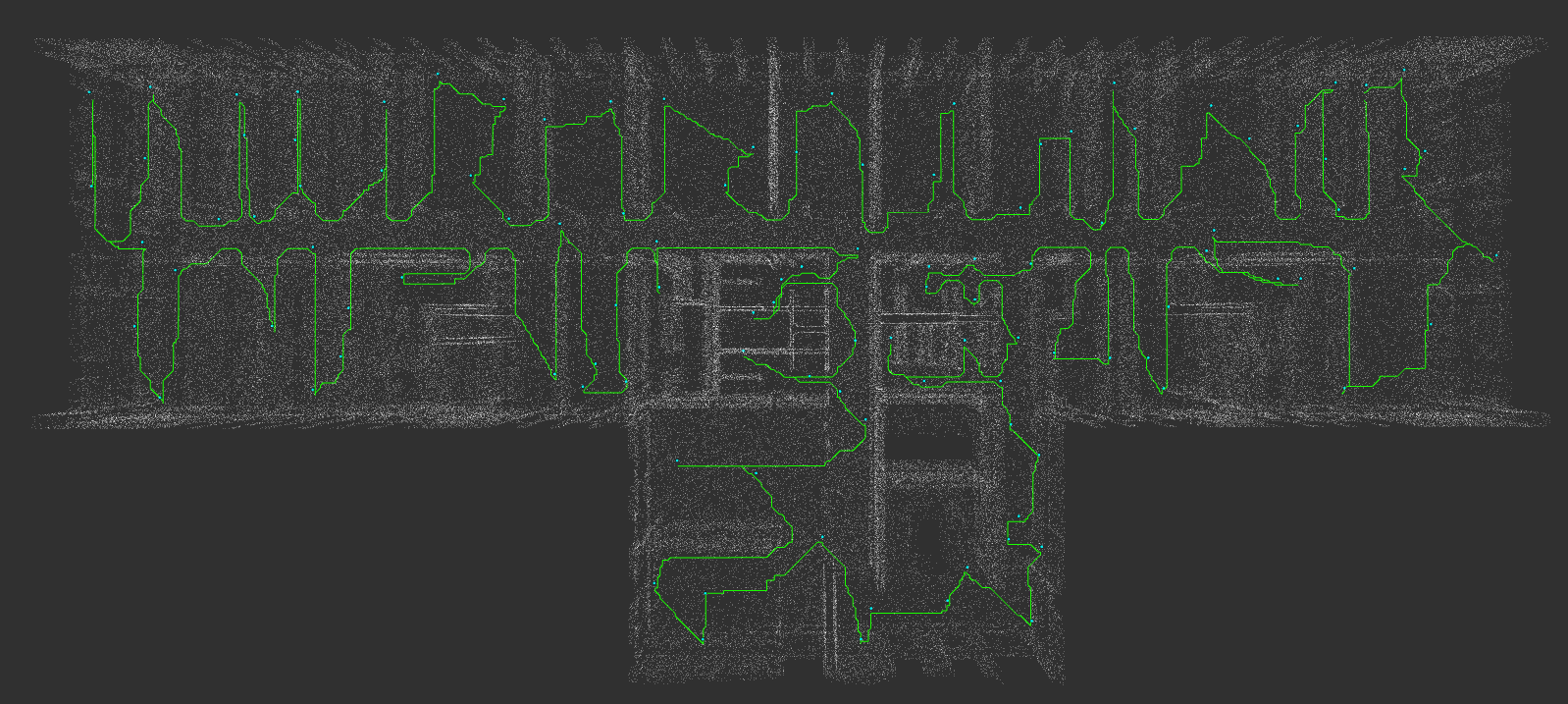}
    \caption{Path planned with the complete model as target model (white point cloud). The dots are the selected viewpoints, the green line is the planned path.}
    \label{fig:precom_res_large_complete}
\end{subfigure}
\caption{Evaluation on real building model using the Spot.}
\label{fig:eval_real_building_model}
\end{figure}

\begin{figure*}[tpb]
\centering
\begin{subfigure}[t]{0.475\linewidth}
  \centering
  \includegraphics[width=\linewidth]{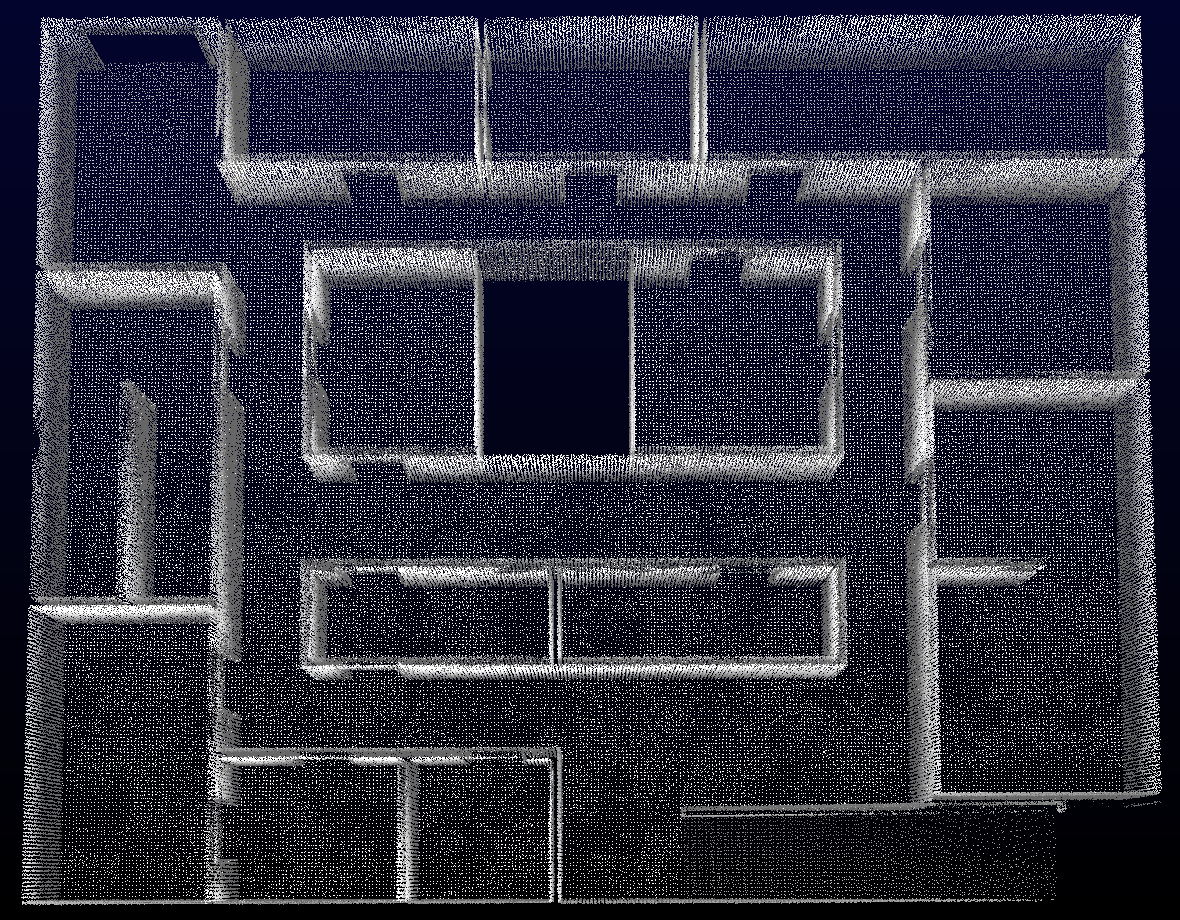}
  \caption{Point cloud recorded with our proposed approach.}
\end{subfigure}\hfill%
\begin{subfigure}[t]{0.475\linewidth}
  \centering
  \includegraphics[width=\linewidth]{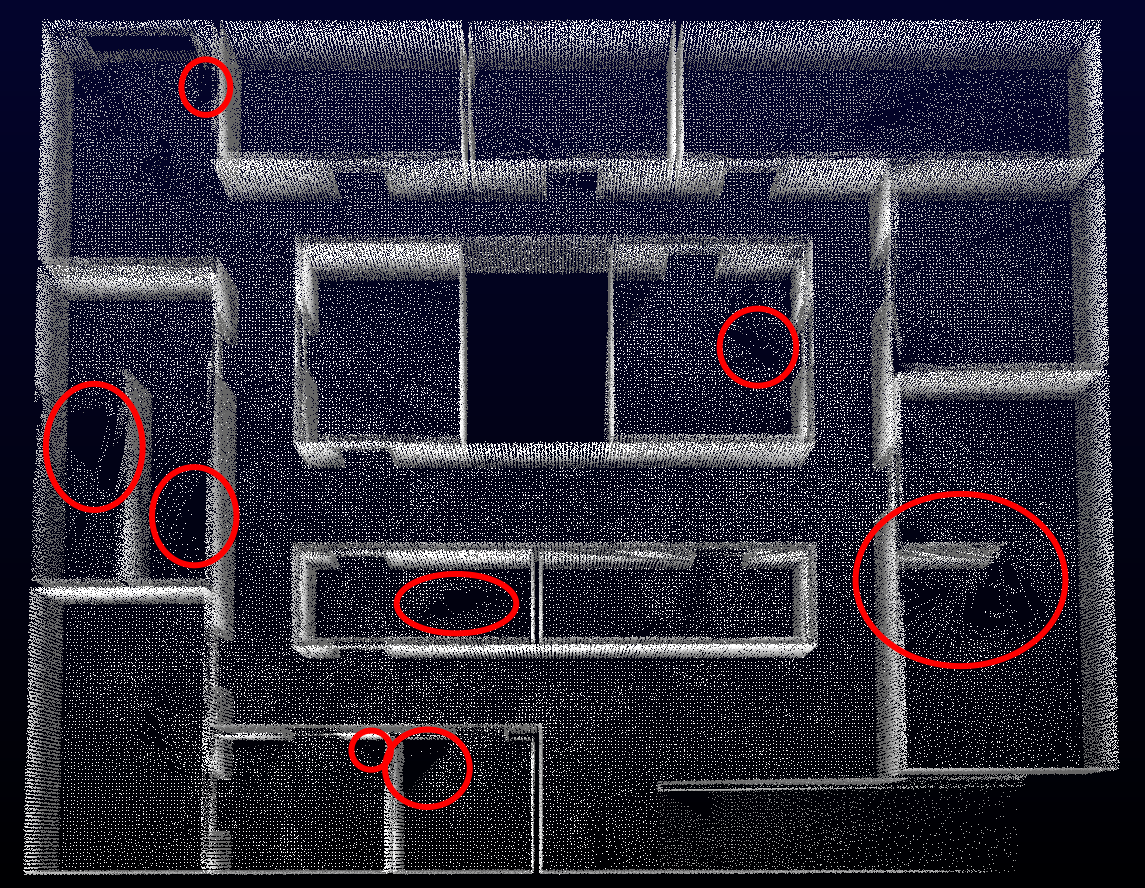}
  \caption{Point cloud recorded during exploration. The highlighted areas are (partially) uncovered.}
\end{subfigure}
\caption{Accumulated point clouds. The evaluations were performed on the artificial model.}
\label{fig:accumulated_point_clouds}
\end{figure*}

On the first floor of the large real building model, driving distances are much greater, so path redundancies have a much greater impact on the total driving distance. This can be seen in the map of the partial exploration using the Spot in Fig. \ref{fig:exploration_path_gridmap_spot_large}. Again, many areas have been left uncovered during the initial traversal of the building. Due to simulation time limitations, the exploration planner was unable to fully map the building. As can be seen in Fig. \ref{fig:precom_res_large_complete} our proposed planner covers all rooms of the building floor without path redundancies. The total path length is \SI{456.3}{\meter} with 111 viewpoints. Additionally, we demonstrate that our approach can handle multiple levels (see Fig. \ref{fig:precom_res_large_target_multilevel}).

Next, we compare the data quality to assess the surface coverage.
During exploration, a point cloud is recorded and accumulated continuously using a voxel filter with a leaf size of \SI{5}{\centi\meter}. In order to make the result of the developed planner comparable, a point cloud was also recorded continuously instead of only at the viewpoints. The results are shown in Fig. \ref{fig:accumulated_point_clouds}. It can be seen that the point cloud of the exploration does have several uncovered areas whereas the accumulated point cloud of our approach is nearly complete.

This comparison shows that our proposed approach can capture the data more effectively by reducing path redundancies while also improving data coverage.

\subsection{Computation time}

Since the proposed approach is an offline planner, the computation time is not a major concern. Additionally, the computation time required for the planning is highly dependent on the used models as well as on chosen parameters. For example, the voxel size of the model representations, the number of target points that are sampled, and the used selector and \acrshort{tsp} solver greatly impact the computation time. However, to give an estimation, some computation times for the planning on the artificial and the real model with the complete model as the target model are shown here. In the evaluation on the artificial model, 327 candidates have been generated, and a \acrshort{sdf} voxel size of \SI{0.0375}{\meter} has been used. The voxel size of the OctoMap was the same for both models, \SI{0.075}{\meter}. For the real building model, a \acrshort{sdf} voxel size of \SI{0.05}{\meter} was chosen and 594 candidates have been generated. The experiments were performed on an \textit{Intel Core i7-8565U CPU @ 1.80GHz}.

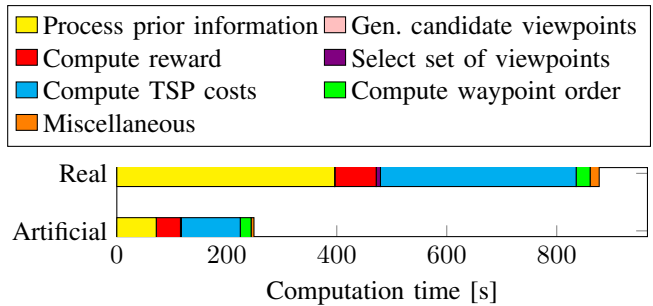
\begin{figure}[thpb]
\begin{tikzpicture}
\begin{axis}[
            height=2.5cm,
            width=\linewidth,
            xbar stacked,
            xmin=0,
            xlabel={Computation time [s]},
            yticklabel style={align=right},
            ytick=data,
            yticklabels from table={\testdata}{Label},
            legend style={at={(0.4,1.3)},anchor=south},
            legend cell align={left},
            legend columns=2,
]
    \addplot [fill=yellow] table [x=prior_info, meta=Label,y expr=\coordindex] {\testdata};
    \addplot [fill=pink] table [x=candidates, meta=Label,y expr=\coordindex] {\testdata};
    \addplot [fill=red] table [x=reward, meta=Label,y expr=\coordindex] {\testdata};
    \addplot [fill=violet] table [x=select_vp, meta=Label,y expr=\coordindex] {\testdata};
    \addplot [fill=cyan] table [x=wp_cost, meta=Label,y expr=\coordindex] {\testdata};
    \addplot [fill=green] table [x=waypoint_order, meta=Label,y expr=\coordindex] {\testdata};
    \addplot [fill=orange] table [x=misc, meta=Label,y expr=\coordindex] {\testdata};
    \legend{Process prior information, Gen. candidate viewpoints, Compute reward, Select set of viewpoints, Compute TSP costs, Compute waypoint order, Miscellaneous}
\end{axis}
\end{tikzpicture}
\caption{Computation time of each step of the planning pipeline for the artificial and real building model.}
\label{fig:computation_time_steps}
\end{figure}

In Fig. \ref{fig:computation_time_steps} the computation time for each step can be seen. The steps have not been parallelized yet, but especially in the rewarding process, an acceleration is to be expected. Also the time for processing the prior information can be reduced significantly by loading previously converted models. However, this only works for multiple planning cycles on the same model with no model parameters changing.

\pgfplotstableread{
Label        prior_info candidates reward select_vp wp_cost waypoint_order misc
Artificial    71.7 0.2 43.8 1.8 107.0 19.7 5.2
Real         396.1 0.3 75.3 7.7 355.9 25.4 16.5
}\testdata

\section{CONCLUSIONS}

We presented a modular time-optimal 3D path planning pipeline for efficient monitoring of construction sites with multiple levels. The use of the 3D building model as prior information enabled the offline planning of paths from which all surfaces of the given 3D model could be covered with a sensor payload. We showed on different building models that paths planned with our approach are much shorter than the ones from existing approaches and provide a higher coverage of the environment.

\bibliographystyle{IEEEtran}
\bibliography{IEEEabrv,bibliography}

\end{document}

%% file: acronyms.tex
\newacronym{POI}{POI}{point of interest}

\newacronym{obj}{.obj}{object file}
\newacronym{ply}{.ply}{Polygon File Format}

\newacronym{pcl}{PCL}{Point Cloud Library}

\newacronym{hdf}{HDF5}{Hierarchical Data Format}

\newacronym{sdf}{SDF}{signed distance field}
\newacronym{tsdf}{TSDF}{truncated signed distance field}
\newacronym{esdf}{ESDF}{Euclidean signed distance field}

\newacronym{bim}{BIM}{building information modeling}
\newacronym{nbv}{NBV}{next best view}

\newacronym{tsp}{TSP}{Traveling Salesman Problem}
\newacronym{mst}{MST}{minimum spanning tree}
\newacronym{sa}{SA}{simulated annealing}

\newacronym{bgl}{BGL}{Boost Graph Library \cite{boost_graph}}

\newacronym{daf}{DAF}{Dynamic Arc Fitting}

\newacronym{ros}{ROS}{Robot Operating System}